\documentclass[11pt,a4paper]{article}
\usepackage[hyperref]{acl2021}
\usepackage{times}
\usepackage{latexsym}
\usepackage[plain]{fancyref}    
\usepackage{float}
\usepackage{graphicx}
\usepackage{tabularx}
\usepackage{cleveref}
\usepackage{multirow}

\usepackage{microtype}

\aclfinalcopy % Uncomment this line for the final submission
\def\aclpaperid{10} %  Enter the acl Paper ID here

\title{Probing Multilingual Language Models for Discourse}
\author{Murathan Kurfalı\\
  Linguistics Department \\
  Stockholm University\\
  Stockholm, Sweden \\
\texttt{murathan.kurfali@ling.su.se } \\\And
Robert Östling\\
  Linguistics Department \\
  Stockholm University\\
  Stockholm, Sweden \\
\texttt{robert@ling.su.se } 
}

\date{}

\begin{document}
\maketitle
\begin{abstract}
Pre-trained multilingual language models have become an important building block in multilingual natural language processing. In the present paper, we investigate a range of such models to find out how well they transfer discourse-level knowledge across languages. This is done with a systematic evaluation on a broader set of discourse-level tasks than has been previously been assembled. We find that the XLM-RoBERTa family of models consistently show the best performance, by simultaneously being good monolingual models and degrading relatively little in a zero-shot setting. Our results also indicate that model distillation may hurt the ability of cross-lingual transfer of sentence representations, while language dissimilarity at most has a modest effect. We hope that our test suite, covering 5 tasks with a total of 22 languages in 10 distinct families, will serve as a useful evaluation platform for multilingual performance at and beyond the sentence level.

\end{abstract}

\section{Introduction}

Large-scale pre-trained neural language models have become immensely popular in the natural language processing (NLP) community in recent years \cite{devlin2019bert,peters2018elmo}. When used as contextual sentence encoders, these models have led to remarkable improvements in performance for a wide range of downstream tasks \cite{qiu2020pre}. In addition, multilingual versions of these models \cite{devlin2019bert,lample2019cross} have been successful in transferring knowledge across languages by providing language-independent sentence encodings.

The general usefulness of pre-trained language models has been convincingly demonstrated thanks to persistent creation and application of evaluation datasets by the NLP community. Discourse-level analysis is particularly interesting to study, given that many of the currently available models are trained with relatively short contexts such as pairs of adjacent sentences.

\newcite{wang2018glue} use a diverse set of natural language understanding (NLU) tasks to investigate the generality of the sentence representations produced by different language models.
\newcite{hu2020xtreme} use a broader set of tasks from across the NLP field to investigate the ability of multilingual models to transfer various types of knowledge across language boundaries.

Our goal in this paper is to systematically evaluate the multilingual performance on NLU tasks, particularly at the discourse level. This combines two of the most challenging aspects of representation learning: multilinguality and discourse-level analysis.
A few datasets have been used for this purpose before, most prominently the XNLI evaluation set \cite{conneau2018xnli} for Natural Language Inference (NLI), and recently also
%PAWS-X \cite{zhang2019paws,yang2019paws} for paraphrase detection as well as
XQuAD \cite{artetxe2019xquad} and MLQA \cite{lewis2019mlqa} for Question Answering (QA). We substantially increase the breadth of our evaluation by adding three additional tasks:
\begin{enumerate}
    \item Penn Discourse TreeBank (PDTB)-style implicit discourse relation classification on annotated TED talk subtitles in seven languages (\Fref{sec:idrc})  
    \item Rhetorical Structure Theory (RST)-style discourse relation classification with a custom set consisting of treebanks in six non-English languages (\Fref{sec:rst})
    \item Stance detection with a custom dataset in five languages
    (\Fref{sec:stance})
\end{enumerate}

We investigate the cross-lingual generalization capabilities of seven multilingual sentence encoders with considerably varying model sizes through their cross-lingual zero-shot performance\footnote{In the remainder of the paper, \textit{cross-lingual zero-shot performance} is simply referred as \textit{zero-shot performance} for brevity. Similarly, source language performance denotes the performance of the respective model on the test set of the training language.} which, in this context, refers to the evaluation scheme where sentence encoders are tested on the languages that they are not exposed to during training. The complied test suite consists of five tasks, covering 22 different languages in total.

We specifically focus on zero-shot transfer scenario where a sufficient amount of annotated data to fine-tune a pre-trained language model is assumed to be available only for one language. We believe that this is the most realistic scenario for a great number of languages; therefore, zero-shot performance is the most direct way of assessing cross-lingual usefulness in a large scale. %Following the same logic, we avoided the translate-train setting, where the training data  for the target languages are created artificially through machine translation, as it obscures the real performance of the models and not directly applicable to many languages due to a lack of decent MT system. 

Our contributions are as follows: (i) we provide a detailed analysis of a wide range of sentence encoders on large number of probing tasks, several of which have not previously been used with multilingual sentence encoders despite their relevancy, (ii) we provide suitably pre-processed versions of these datasets to be used as a multilingual benchmark for future work with strong baselines provided by our evaluation, (iii) we show that the zero-shot performance on discourse level tasks are not correlated with any kind of language similarity and hard to predict, (iv) we show that knowledge distillation may selectively destroy multilingual transfer ability in a way that harms zero-shot transfer, but is not visible during evaluations where the models are trained and evaluated with the same language.

% we have, in addition to the XTREME, distilbert,2 XLM variations and XLM-R-Base which I believe will let us say something about the relation between the model size, training time and performance.

\section{Background} \label{sec:bg}
The standard way of training a multilingual language model is through a large non-parallel multilingual corpora, e.g.\ Wikipedia articles, where the models are not provided with any explicit mapping across languages which renders cross-lingual performance of such models puzzling. \newcite{pires2019multilingual} and \newcite{wu2019beto} are the earliest studies to explore that puzzle by trying to uncover the factors that give multilingual BERT (henceforth, mBERT) its cross-lingual capabilities. \newcite{pires2019multilingual} perform a number of probing tasks and hypothesize that the shared sentence pieces across languages gives mBERT its generalization ability by forcing other pieces to be mapped into the same space. Similarly, \newcite{wu2019beto} evaluate the performance of mBERT in five tasks and report that while mBERT shows a strong zero-shot performance, it also retains language-specific information in each layer.

%link: https://arxiv.org/pdf/1909.00142.pdf
% maybe mention later: Arxiv paper focusing on discourse-based evaluation: https://arxiv.org/pdf/1907.08672.pdf . 

\newcite{chen2019evaluation} proposes a benchmark to evaluate sentence encoders specifically on discourse level tasks. The proposed benchmark consists of discourse relation classification and a number of custom tasks such as finding the correct position of a randomly moved sentence in a paragraph or determining if a given paragraph is coherent or not. The benchmark is confined to English, hence, only targets monolingual English models. 

% google's paper: https://arxiv.org/pdf/2003.11080.pdf
% X-Glue paper: https://arxiv.org/pdf/2004.01401.pdf

%% NEW PAPER: https://arxiv.org/pdf/2005.00633.pdf
% they evaluated BERT, XLM-R on (POS , dependency parsing, NER) (NLI, QA) and show that few-shot learning is way better than zero shot. they do cool analysis, check correlation between the size of the training data / language similarities and their zero shot performance; they take language similarities from Lang2vec. I added such a correlation analysis to our discussion as well (last section).

Two very recent studies, XTREME \cite{hu2020xtreme} and XGLUE \cite{liang2020xglue},  constitute the first studies on the cross-lingual generalization abilities of pre-trained language models via their zero-shot performance. The tasks in both studies largely overlap, where XTREME serves as cross-lingual benchmark consisting of well-known datasets, e.g. XNLI, XQuAD. On the other hand, while covering the most of XTREME tasks\footnote{Except parallel sentence retrieval tasks.}, XGLUE offers new datasets which either focus on the relation between a pair of inputs, such as web page--query matching, or on text generation via question/news title generation. In addition to the mBERT and certain XLM and XLM-R versions, XTREME includes MMTE \cite{arivazhagan2019massively} whereas XGLUE evaluates Unicoder \cite{huang2019unicoder} among its baselines.

\section{Cross-lingual Discourse-level Evaluation}

In discourse research, sentences/clauses are not understood in isolation but in relation to one another. The semantic interactions between these units are usually regarded as the backbone of coherence in various prominent discourse theories including that underlying the Penn Discourse TreeBank (PDTB) \cite{prasad2007penn}, and Rhetorical Structure Theory (RST) \cite{MannThompson1988rst} used in the RST Discourse Treebank \cite{carlson2001discourse}. Modelling such interactions requires an understanding that is beyond sentence-level and, from this point-of-view, determining any kind of relation between sentences/clauses can be associated with discourse.

Although paraphrase detection or natural language inference may not strike as discourse-level tasks at first glance, they both deal with semantic relations between sentences. \newcite{tonelli2012hunting} show that textual entailment is, in fact, a subclass of \textit{Restatement} relations of the PDTB framework whereas \newcite{nie2019dissent} report an increase in discourse relation classification accuracy when NLI is used as the intermediate fine-tuning task. In a similar vein, a stance against a judgement, \textit{Favor} or \textit{Against}, can be seen as \textit{CONTINGENCY: Cause: reason} and \textit{COMPARISON: Contrast} in PDTB; \textit{Explanation} and \textit{Antithesis} in RST, respectively. 

Therefore, these NLU tasks can be seen as special subsets of discourse relation classification; only a model with a good understanding beyond individual sentences can be expected to solve these tasks. Finally, since question answering requires an understanding on discourse level in order to be solved, so we also believe classifying this as a discourse-level task should be uncontroversial.

\subsection{Tasks \& Datasets}
%% TODO: if page limit permits add table with an example training instance from each task

%A discourse relation mark the relation between discourse units which is beyond the meaning conveyed by the individual units. 
In this section, we present our task suite and the datasets used for training and zero-shot evaluation. For the sake of clarity, we name each task after the dataset used for training. 

%\textbf{Implicit Discourse Relation Classification (PDTB): }
\subsubsection{Implicit Discourse Relation Classification (PDTB)}
\label{sec:idrc}
Implicit discourse relations hold between adjacent sentence pairs but are not explicitly signaled with a connective such as \textit{because}, \textit{however}. Implicit discourse relation classification is the task of determining the sense conveyed by these adjacent sentences, which can be easily inferred by readers. Classifying implicit relations constitutes the most challenging step of shallow discourse parsing \cite{xue2016conll}. 

The training is performed on PDTB3 \cite{webber2016discourse} where sections 2--20, 0--1 are used for training and development respectively. The zero-shot evaluation is performed on the TED-MDB corpus \cite{zeyrek2019ted}\footnote{https://github.com/MurathanKurfali/Ted-MDB-Annotation}, which is a PDTB-style annotated parallel corpus consisting of 6 TED talk transcripts, and the recent Chinese annotation effort on TED talk transcripts that however are mostly not parallel to TED-MDB \cite{long2020shallow}. Due to the small size of the test sets, we confine ourselves to the top-level senses: \textit{Contingency, Comparison, Expansion, Temporal} which is also the most common setting for this task. Despite the limited size of TED-MDB, zero-shot transfer is possible and yields meaningful results as shown in \cite{kurfali2019zero}. In total, seven languages are evaluated in this task: English, German, Lithuanian\footnote{Lithuanian is the latest addition to the Ted-MDB corpus, as documented in \cite{oleskeviciene2018observations}.}, Portuguese, Polish, Russian and Chinese.

\subsubsection{Rhetorical Relation Classification (RST)}
\label{sec:rst}
%\textbf{Rhetorical Relation Classification (RST)}:
Rhetorical relations are just another name for discourse relations but this term is most commonly associated with Rhetorical Structure Theory (RST) \citep{MannThompson1988rst}. Similar to PDTB's discourse relations, rhetorical relations also denote links between discourse units, but are considerably different from the former. The difference largely stems from the take of the respective theories on the structure of the discourse. RST conceives discourse as one connected tree-shaped structure assuming hierarchical relations among the discourse relations. On the other hand, PDTB does not make any claims regarding the structure of the discourse and annotates discourse relations only in a local context (i.e.\ adjacent clauses/sentences) without assuming any relation on higher levels. Hence, evaluation on RST and PDTB relations can be seen as complementary to each other as the former focuses on both global and local discourse structure whereas PDTB focuses only on local structure.

We use English RST-DT \cite{carlson2001discourse} for training where a randomly selected 35 documents are reserved for development. However,
unlike PDTB, there is not any compact parallel RST corpus; RST annotations across languages usually differ from each other in several ways. Therefore, we follow \newcite{braud2017cross} and create a custom multilingual corpus for the zero-shot experiments which consists of the following languages: Basque \cite{iruskieta2013rst}, Brazilian Portuguese \cite{cardoso2011cstnews,collovini2007summ,pardo2005rhetalho}, Chinese \cite{cao2018rst}, German \cite{stede2004potsdam}, Spanish \cite{da2011development},  Russian \cite{pisarevskaya2017towards}. We perform a normalization step on each treebank which includes binarization of non-binary trees and mapping all relations to 18 coarse grained classes described in \cite{carlson2001discourse}. The normalization step is performed via the pre-processing scripts of \cite{braud2017cross}. Due to memory constraints, we limit the sequence lengths to 384. Hence, we only keep those relations where the first discourse unit is shorter than 150 words so that both units can be equally represented which lead to omission of only 5\% of all non-English relations.

\subsubsection{Stance Detection (X-Stance)}
\label{sec:stance}
%\paragraph{\textbf{Stance Detection (X-Stance): }}
The stance detection is task of determining the attitude expressed in a text towards a target claim. For experiments, we mainly use the X-stance corpus which consists of 60K answers to 150 questions concerning politics in German, Italian and French \cite{vamvas2020x}. Unlike other tasks, we select German as the training language for stance detection as it is the largest language in X-Stance. Following the official split, we use the German instances in the training and development sets during fine-tuning and non-German instances in the test set for evaluation. Furthermore, we enrich the scope of our zero-shot evaluation by two additional dataset, one in English \cite{chen2019seeing} and other one in Chinese \cite{yuan2019exploring}, which also consist of stance annotated claim--answer pairs, despite in different domains. 

\subsubsection{Natural Language Inference (XNLI)}
\label{sec:nli}
%\paragraph{\textbf{Natural Language Inference (XNLI): }}
Natural language inference (NLI) is the task of determining whether a premise sentence entails, contradicts or is neutral to a hypothesis sentence. MultiNLI and the mismatched part of the development data \cite{williams2017broad} are used for training and validation, respectively. The evaluation is performed on the test sets of the XNLI \cite{conneau2018xnli} corpus which covers the following 14 languages in addition to English: French, Spanish, German, Greek, Bulgarian, Russian, Turkish, Arabic, Vietnamese, Thai, Chinese, Hindi, Swahili and Urdu.

%\subsubsection{Paraphrase Detection (PAWS-X)}
%\label{sec:paraphrase}
%Paraphrase detection refers to the task of determining whether two sentences are paraphrases of each other. The English PAWS-X \cite{zhang2019paws} data is used for training whereas the test sets are used for evaluation. 

\subsubsection{Question Answering (XQuAD)}
\label{sec:qa}
%\paragraph{\textbf{Question Answering (Xquad): }}
Question answering is the task of identifying span in a paragraph which answers to a question. We use the SQuAD v1.1 \cite{rajpurkar2016squad} for training. We evaluate the models on the popular XQuAD dataset which contains the translation of SQuAD v1.1 development set into ten languages \cite{artetxe2019xquad}: Spanish, German, Greek, Russian, Turkish, Arabic, Vietnamese, Thai, Chinese, and Hindi.

\begin{table*}[t]
\centering
\begin{tabular}{l|ccccc}
Task & Training data & $|train|$  &$|test|$ &\#langs&metric\\ \hline
RST & RST DT & 17K & 603 -- 6,902 &6 & acc \\
PDTB & PDTB3 & 17K & 194 -- 1,366 &7& F$_1$ \\
X-stance& X-stance-DE& 33K &1,446 -- 6,153 &4& F$_1$\\
NLI & MultiNLI & 433K & 5,010 &14 & acc\\
%PAWS  & PAWS-EN&  43K& 2,000 &6&acc\\
%MLDoc & MLDoc-EN &10K &4,000 (7)& acc\\
Q/A & Squad 1.1 & 100K & 1,190 &11& ex. match/F$_1$\\
\end{tabular}
\caption{Summary of the datasets used in experiments. "Corpus name-(lang.code)" refers to the part of the corpus belonging to the respective language. \#langs refers to the number of \emph{zero-shot} languages, excluding the training language.}
\label{tab:summary}
\end{table*}

\subsection{Languages}

The proposed task suite covers the following 22 languages representing 10 language families:
Indo-European (Bulgarian \textit{bg},  German \textit{de}, Greek \textit{el}, English \textit{en}, Spanish \textit{es}, French \textit{fr}, Hindi \textit{hi}, Italian \textit{it}, Lithuanian \textit{lt}, Polish \textit{pl}, Portuguese \textit{pt}, Russian \textit{ru}, Urdu \textit{ur}),  Afroasiatic (Arabic \textit{ar}), Basque (\textit{eu}), Japonic (Japanese \textit{ja}), Koreanic (Korean \textit{ko}), Niger-Congo (Swahili \textit{sw}), Tai-Kadai (Thai \textit{th}), Turkic (Turkish \textit{tr}), Austroasiatic (Vietnamese \textit{vi}), Sino-Tibetan (Chinese \textit{zh}). 
%Indo-European (\textit{bg, de, el, en, es, fr, hi, it, lt, pl, pt, ru, ur}), Afroasiatic (\texit{ar}), Basque (\textit{eu}), Japonic (\textit{ja}), Koreanic (\textit{ko}), Niger-Congo (\textit{sw}), Tai-Kadai (\textit{th}), Turkic (\textit{tr}), Austroasiatic (\textit{vi}), Sino-Tibetan (\textit{zh}). 
%\textit{ar, bg, de, el, en, es, eu, fr, hi, it, ja, ko, lt, pl, pt, ru, sw, th, tr, ur, vi, zh}.
Seven of these languages are evaluated in at least three different tasks. 

\section{Experiments}

We evaluate a wide range of multilingual sentence encoders which learn contextual representations. The evaluated models represent a broad spectrum of model sizes, in order to allow practitioners to estimate the trade-off between model size and accuracy.

\subsection{Sentence Encoders} \label{sec:lm}

The sentence encoders evaluated in the current paper are described in detailed below, and their characteristics summarized in 
\Fref{tab:lms}.

\paragraph{Multilingual BERT (mBERT):} mBERT is a transformer-based language model trained with masked language modelling and next sentence prediction objectives similar to the original English BERT model \cite{devlin2019bert}\footnote{\url{https://github.com/google-research/bert/blob/master/multilingual.md}}. mBERT is pre-trained on the Wikipedias of 104 languages with a shared word piece vocabulary. As discussed in Section \ref{sec:bg}, its input is not marked with any language-specific signal and mBERT does not have any objective to encode different languages in the same space. 

\paragraph{distilmBERT:} distilmBERT is a compressed version of mBERT obtained via model distillation \cite{sanh2019distilbert}. Model distillation is a compression technique where a smaller model, called \textit{student}, learns to mimic the behavior of the larger model, called \textit{teacher}, by matching its output distribution. distilmBERT is claimed to reach 92\% of mBERT's performance on XNLI while being two times faster and 25\% smaller.\footnote{\url{https://github.com/huggingface/transformers/tree/master/examples/distillation}} However, to the best of our knowledge, there is not any comprehensive analysis of distilmBERT's zero-shot performance.

\paragraph{XLM:} XLM is a transformer-based language model aimed at extending BERT to cross-lingual setting \cite{lample2019cross}. To this end, XLM increases the shared vocabulary across languages via shared byte pair encoding (BPE) vocabulary. Moreover, unlike BERT, the input sentences are accompanied by language embeddings. There are several different XLM models which differ at either number of training languages or training objectives. In the current study, we consider the following three:

\begin{itemize}
    \item \textit{XLM-mlm}: The XLM model which is trained with BERT's masked language model (MLM) objective on the Wikipedias of the 15 XNLI languages.
    
    \item \textit{XLM-tlm}: In addition to the MLM, this XLM model has a novel training objective which is called Translation Language Model (TLM). In TLM, the model receives a pair of translationally equivalent sentences and tries to predict the masked word by attending both sentences. Hence, the model tries to predict the masked word by looking at its context in another language which encourages representations of different languages to be aligned. TLM is shown to lead a significant increase on XNLI \cite{lample2019cross}. XLM-tlm is also trained for 15 XNLI languages but only on parallel data.
    
    \item \textit{XLM-100}: This version is trained, like mBERT, on Wikipedia data covering 100 languages using only an MLM objective. Unlike previous XLM models, this version does not utilize language embeddings. 
\end{itemize}

\paragraph{XLM-RoBERTa (XLM-R):} XLM-RoBERTa is not an XLM model, in spite of what its name suggests. XLM-R does not use language embeddings, applies sentence-piece tokenization instead of BPE and is not trained on a parallel corpus unlike the XLM-tlm. Instead, it is a RoBERTa model \cite{liu2019roberta}, which is an optimized version of BERT, trained on 2.5 TB of cleaned CommonCrawl data covering 100 languages \cite{conneau2019unsupervised}. There are two released XLM-R models, XLM-R$_{base}$ and XLM-R$_{large}$, named after the BERT-architecture they are based on. Compared to original multilingual-BERT, XLM-RoBERTa models have a considerably larger vocabulary size which results in larger models.

\begin{table*}[t]
\centering
\begin{tabular}{l|cccc}
Model&Langs& Parameter count& Vocab. size & \# of layers  \\
\hline
distilmBERT&104&134M&30K&6 \\ \hline
mBERT&104&177M&30K&12 \\ \hline
XLM-mlm &15&250M&95K&12  \\
XLM-tlm &15&250M&95K&12  \\
XLM-100 &100&570M&200K&16 \\ \hline
XLM-R$_{base}$&100&270M&250K&12 \\
XLM-R$_{large}$&100&550M&250K&24 \\
\end{tabular}
\caption{The characteristics of the sentence encoders evaluated in the experiments}
\label{tab:lms}
\end{table*}

\subsection{Experimental Setup}

A summary of the datasets used in the experiments is provided in Table \ref{tab:summary}. Except PDTB, all datasets are publicly available. As stated earlier, the training language is English for all tasks except stance detection where German is preferred due the size of the available data. In the spirit of real zero-shot transfer, the validation sets only consist of instances in the training language; hence, no cross-lingual information whatsoever is utilized during training/model selection. For the evaluation metrics, we stick to the default metrics of each task (Table \ref{tab:summary}).

We set the sequence length to 384 for question answering and RST relation classification; to 250 for stance detection and to 128 for the remaining tasks. At evaluation time, we keep the same configuration. For all models, adam epsilon is set to 1e-8 and maximum gradient norm to 1.0. The learning rate of $2 \times 10^{-5}$ is used for all the models except XLM-R-large and XLM-100 where it is set to $5 \times 10^{-6}$. We adopt the standard fine-tuning approach and fine-tune all models for 4 epochs. We do not apply any early stopping and use the model with the best validation performance during zero-shot experiments. All tasks are implemented using Huggingface's Transformers library \cite{wolf2019huggingface}. As fine-tuning procedure is known to show high variance on small training datasets, all models are run for 4 times with different seeds and the average performance is reported. For XLM and XLM-tlm models, we fall back to English language embeddings for non-XNLI languages. All experiments are run on a single TITAN X (12 GB) GPU.

\section{Results and Discussion}

We provide an overview of the main results in Figure \ref{fig:results}. The detailed results with per-language breakdown are provided in the Appendix \ref{sec:appendix}. 

Overall, there is a clear difference between the training and zero-shot performance of all models. When averaged over all tasks, the performance loss in zero-shot transfer ranges from 15.58\% (XLM-R-large) to 34.96\% (distilmBERT) which clearly highlights the room for improvement, especially with smaller model sizes. In the rest of the section, we discuss the results in terms of the encoder type, task and the languages.

\begin{figure}[t]{}
    \centering
\includegraphics[width=0.37\paperwidth]{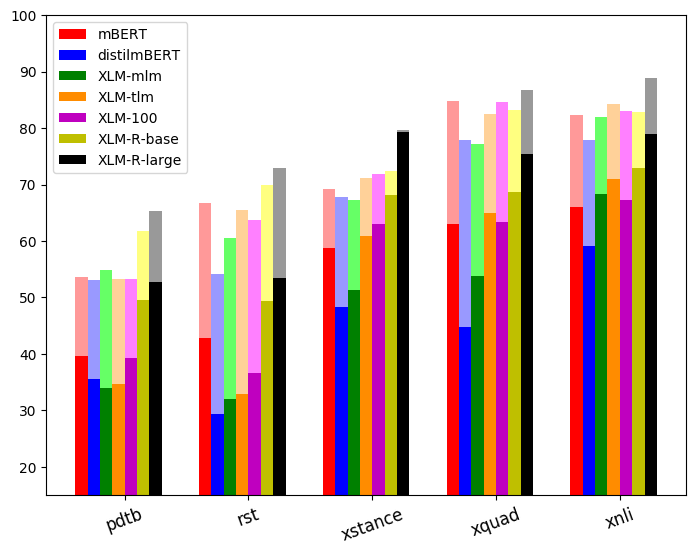}    
\caption{Overview of performance of each sentence encoder on all Disco-X tasks. The semi-transparent bars represent source language performance (German for X-stance, English for the rest) while the solid bars represent the zero-shot performance, i.e.\ the mean performance across all languages \emph{except} the training language. All values are averages over independent training runs.}
\label{fig:results}
\end{figure}

\paragraph{Model-wise analysis} 

The ranking of the encoders displays relatively little variation across tasks, with XLM-R$_{large}$ exhibiting the best zero-shot performance across all tasks by outperforming the second best model (XLM-R$_{base}$) by 5.98\%. distillmBERT, on the other hand, fails to match the performance of other encoders.\footnote{The only exception is the XLM and XLM-tlm's performance on non-XNLI languages where distillmBERT manages to outperform them but not always by a large margin.}

The Translation Language Model (TLM) objective is proved to be a better training objective than MLM by consistently outperforming the vanilla XLM in all tasks. XLM-tlm outperforms XLM-100 on XNLI languages as well which is possibly because of the \lq curse of multilinguality\rq\ \cite{conneau2019unsupervised}, the degradation of the overall performance in proportion to the number of languages in the training. However, training setting (e.g.\ training data, hyperparameters) outplays the \lq curse of multilinguality\rq\ as XLM-R$_{base}$ clearly outperforms XLM-tlm even on XNLI languages. It would be interesting to see how an XLM-R trained with TLM objective on small set of languages, e.g. XNLI languages, would perform.

DistillmBERT is the lightest model evaluated in the current investigation. It is shown to retain 92\% of the mBERT's performance on certain XNLI languages.\footnote{\url{https://github.com/huggingface/transformers/tree/master/examples/distillation}} The results suggest that distillmBERT delivers its promise, although to a lesser extent. When averaged over all tasks, distillmBERT retains 93\% of the source language performance of mBERT. However, its relative performance significantly drops to 82\% on zero-shot transfer. That is, distillmBERT is not as successful when it comes to copying mBERT's cross-lingual abilities. Furthermore, its performance (relative to mBERT) is not stable across tasks either. It only achieves 69\% of mBERT's zero-shot performance on RST  whereas 89\% on XNLI. The low memory requirement and its speed (with the same batch size, it is x2 faster than mBERT and x5 than XLM-R$_{large}$) definitely makes distillmBERT a favorable option; however, the results show that its zero-shot performance is considerably lower than its source language performance and is highly task-dependent, hence, hard to predict.

\begin{table*}[]
\centering
\begin{tabular}{l|llllll}
Model &   PDTB &    RST & X-stance &  XQuAD &   MNLI &      Average $ \pm$ std \\ \hline
mBERT       &  74.49 &  64.18 &   84.75 &  74.22 &  80.28 &   $75.58 \pm 6.92$ \\
distilmBERT &  66.13 &  54.37 &   71.34 &  57.35 &   75.9 &   $65.02 \pm 8.15$ \\
XLM-mlm     &  60.32 &  52.93 &    76.4 &  69.68 &  83.47 &  $68.56 \pm 10.93$ \\
XLM-tlm     &  63.49 &  50.36 &   85.57 &  78.76 &  84.26 &  $72.49 \pm 13.56$ \\
XLM-100     &  73.76 &  57.54 &   87.62 &  74.89 &  81.01 &  $74.96 \pm 10.02$ \\
XLM-R$_{base}$  &  78.96 &  70.75 &   94.29 &  82.44 &   88.1 &   $82.91 \pm 8.00$ \\
XLM-R$_{large}$ &  79.91 &  73.33 &   100.4 &  86.81 &     89 &   $85.89 \pm 9.11$ \\ \hline
%Average     &  71.01 &  60.49 &   85.77 &  74.88 &  83.15 &   $75.06 \pm 9.04$ \\
\end{tabular}
\caption{Relative zero-shot performance of each encoder to the source language performance (metrics differ between tasks but higher is better in all cases). The figures shows what percentage of the source language performance is retained through zero-shot transfer in each task. \newcite{hu2020xtreme} refer to this as the \emph{cross-lingual transfer gap}. A score above 100 indicates that a better zero-shot performance than that of training.}
\label{tab:zero}
\end{table*}

\paragraph{Task-wise Analysis}

%Despite all being discourse-level tasks, the performance of the models vary considerably across them. 

%We find that the discourse relation classification tasks (both RST and PDTB) are the most challenging tasks followed by question answering. When the complexity of tasks are considered, it is quite intuitive; discourse relation classification tasks are multi-way classification tasks where RST has as many as 18 labels whereas other tasks are either two-way (X-stance) or three-way (XNLI). Moreover, none of the resources that form PDTB and RST datasets are crowd-sourced but annotated by expert annotators. Hence, they are less likely to be prone to annotation artifacts, as is the case with some large-scale datasets including MNLI \cite{gururangan2018annotation}, which is also partly suggested by the low performances.

Table \ref{tab:zero} shows to what extent encoders manage to transfer their source language performance to zero-shot languages. Overall, the zero-shot performances show high variance across tasks which is quite interesting given that all tasks are on the same linguistic level. It is also surprising that mBERT manages a better zero-shot transfer performance than all XLM models while being almost as consistent as XLM-R$_{base}$.  

Overall, the results show that even modern sentence encoders struggle to capture inter-sentential interactions in both monolingual and multilingual settings, contrary to the what the high performances on well-known datasets (e.g.\ PAWS \cite{hu2020xtreme}) may suggest. We believe that this finding supports our motivation to propose new probing tasks to have a fuller picture of the capabilities of these encoders.

\paragraph{Language-wise Analysis:}  In all tasks, regardless of the model, training-language performance is better than even the best zero-shot performance. The only exception is the XLM-R-large's performance on the X-stance where the zero-shot performance is on par with its performance on the German test set. 

An important aspect of cross-lingual research is predictability. The zero-shot performance of a certain language do not seem to be stable across tasks (e.g. German is the language with the worst RST performance; yet it is one of the best in XNLI). We further investigate this following \newcite{lauscher2020zero}, who report high correlation between syntactic similarity and zero-shot performance for low-level tasks, POS-tagging and dependency parsing. We conduct the same correlation analysis using Lang2Vec \cite{littell2017uriel}. However, syntactic and geographical similarity only weakly correlates with zero-shot performances across the tasks (Pearson's $r = .46$ and Spearman's $r = .53$  on average for syntactic; Pearson's $r = .30$ and Spearman's $r = .45$ for geographical similarity). Such low correlations are important as it further supports the claim that the tasks are beyond the sentence level and also highlights a need for further research to reveal the factors at play during zero-shot transfer of discourse-level tasks.

%Interestingly, the zero-shot performance on the PAWS dataset highly correlates with syntactic similarity to the training language (English) across all encoders. We believe that this is the case because the PAWS dataset consists of pairs created by back-translation and word-swapping. Therefore, the task can be solved without having to model higher-level semantic interaction between sentences and represent only an aspect of paraphrase detection problem. 

%%% from the PAWS-X paper: "Our experimental results showed that PAWS-X effectively measures sensitivity of models to word order and the efficacy of cross-lingual learning approaches." so they admit it measures word-order more than the semantics.

% Robert: will start with the abstract, which could then be partly plagiarized here.

\section{Conclusion}
%% I commented out this: \footnote{The resources will be released upon acceptance.} .
As pre-trained multilingual sentence encoders have become prevalent in natural language processing, research on cross-lingual zero-shot transfer gains increasing importance \cite{hu2020xtreme,liang2020xglue}. In this work, we evaluate a wide range of sentence encoders on a variety of discourse-level tasks in a zero-shot transfer setting. Firstly, we enrich the set of available probing tasks by introducing three resources which have not been utilized in this context before. We systematically evaluate a broad range of widely used sentence encoders with considerably varying sizes, an analysis which has not been made before.

The main variable we look at is the performance gap between training-language evaluation and zero-shot evaluation. Unsurprisingly, nearly always there is such a gap, but its magnitude depends on a number of factors:
\begin{itemize}
    \item \textbf{Distillation}: the distilled mBERT model has a larger gap than the full mBERT model, indicating loss of multilingual transfer ability during distillation.
    \item \textbf{Language similarity}: the gap correlates only weakly with measures of language similarity (syntactic and geographical), indicating that sentence encoders generally transfer discourse-level information about as well between similar and dissimilar languages.
    \item \textbf{High variance}: apart from the above, we also observe a generally high variance in the gap magnitude between different tasks in our benchmark suite.
\end{itemize}

These observation provide several starting points for future work: investigating why knowledge distillation seems to hurt zero-shot performance to a much greater extent than same-language sentence encoding ability and what can be done to solve this problem, and explaining the large variations in the zero-shot transfer gap between different discourse-level NLP tasks.

%The results show that none of the models manage to retain their monolingual performance during zero-shot transfer, and interestingly enough the difference is especially large for the distillation based model.

%Moreover, the performance loss during zero-shot performance is hard to predict suggested by (i) the difference in performance across tasks despite all of them being on the same linguistic level, (ii) the low correlations between language similarity and zero-shot performance. The results suggests that there is plenty of room for improvement in both transferring the discourse-level knowledge across languages and preserving cross-lingual abilities with distilled models. 

\bibliography{anthology,eacl2021}
\bibliographystyle{acl_natbib}

\appendix

\newpage
\section{Task-wise Results} \label{sec:appendix}
\noindent\begin{minipage}{\textwidth}
\centering
\begin{tabular}{l|c|ccccccc }
Model&en&de&es&eu&pt&ru&zh&AVG\\ \hline
mBERT&66.7&29.2&39.3&31.1&58.6&48.0&50.7&42.8\\
distilmBERT&54.1&16.3&25.7&21.4&44.5&32.2&36.5&29.4\\
XLM-mlm&60.6&25.5&33.2&14.1*&40.4*&39.8&39.4&32.1\\
XLM-tlm&65.5&26.0&35.2&13.3*&42.0*&39.9&41.3&33.0\\
XLM-100&63.8&24.3&34.6&26.2&55.2&40.0&39.8&36.7\\
XLMR-b&69.8&37.6&44.7&39.4&61.9&56.2&56.7&49.4\\
XLMR-l&72.9&44.8&46.8&47.0&65.6&59.3&57.3&53.5\\
\end{tabular}
\captionof{table}{RST zero-shot results (Accuracy) for each language. * denotes that the language is not one of the training languages of the respective sentence encoder.} \label{tab:rst}
\end{minipage}
\noindent\begin{minipage}{\textwidth}
\centering
\begin{tabular}{l|c|cccccccc }
Model&en&de&lt&pl&pt&ru&tr&zh&AVG\\ \hline
mBERT&53.6&42.7&39.2&33.9&46.7&33.1&40.3&43.5&39.9\\
distilmBERT&53.1&42.7&30.0&34.7&41.1&32.6&29.4&35.4&35.1\\
XLM-mlm&54.9&44.9&19.5*&20.6*&28.9*&33.8&43.5&40.5&33.1\\
XLM-tlm&53.3&45.9&20.1*&21.3*&26.8*&37.1&41.9&43.6&33.8\\
XLM-100&54.6&41.9&41.6&32.5&44.5&34.2&35.9&40.4&38.7\\
XLMR-b&61.8&49.5&49.6&40.4&53.5&42.7&54.4&51.4&48.8\\
XLMR-l&65.4&53.4&49.4&42.8&59.5&48.9&53.8&58.1&52.3\\
\end{tabular}
\captionof{table}{PDTB zero-shot results (F$_1$) for each language. * denotes that the language is not one of the training languages of the respective sentence encoder.}\label{tab:pdtb}
\end{minipage}
\noindent\begin{minipage}{\textwidth}
\centering
\begin{tabular}{l|c|ccccc }
Model&de&en&fr&it&zh&AVG\\ \hline
mBERT&69.3&60.2&60.7&63.2&50.8&58.7\\
distilmBERT&67.7&49.8&48.7&59.5&35.2&48.3\\
XLM-mlm&67.3&52.6&55.0&56.2*&41.8&51.4\\
XLM-tlm&71.2&60.4&62.5&59.6*&61.1&60.9\\
XLM-100&71.8&62.3&64.8&64.0&60.6&62.9\\
XLMR-b&72.3&65.8&70.4&69.9&66.7&68.2\\
XLMR-l&79.3&80.9&79.0&78.9&79.5&79.6\\
\end{tabular}
\captionof{table}{X-stance zero-shot results (F$_1$) for each language. * denotes that the language is not one of the training languages of the respective sentence encoder.}
\label{tab:xstance}
\end{minipage}
\noindent\begin{minipage}{\textwidth}
\centering
\small
\begin{tabular}{p{42pt}|p{14pt}|p{14pt}  p{13pt}   p{13pt}  p{13pt}  p{13pt}  p{13pt}  p{13pt}  p{13pt}  p{13pt}  p{13pt}  p{13pt}  p{13pt}  p{13pt}  p{13pt} p{13pt} }
Model&en&ar&bg&de&el&es&fr&hi&ru&sw&th&tr&ur&vi&zh&AVG\\ \hline
mBERT&82.3&65.7&69.4&72.1&68.2&75.9&75.3&60.6&69.8&51.3&54.7&62.2&58.8&70.9&69.7&66.1\\
distilmBERT&77.9&60.3&63.9&65.7&61.4&70.1&69.9&54.7&63.6&46.6&39.1&57.3&54.1&59.2&62.4&59.2\\
XLM-mlm&81.9&68.5&73.7&73.0&73.3&75.3&75.2&64.4&72.0&64.9&49.2&67.3&62.8&70.3&67.3&68.4\\
XLM-tlm&84.2&71.1&76.5&76.2&74.3&78.3&77.9&66.5&75.3&67.4&53.9&70.8&62.7&72.8&69.3&70.9\\
XLM-100&83.1&67.9&72.6&73.3&72.4&76.6&75.5&64.7&71.3&58.4&39.7&68.2&62.0&72.7&67.0&67.3\\
XLMR-b&82.8&71.0&77.3&75.7&75.3&78.2&76.9&68.6&75.2&66.4&71.6&72.4&65.2&74.6&73.0&73.0\\
XLMR-l&88.8&78.6&83.0&82.9&81.8&84.5&82.7&76.0&79.3&71.6&77.0&78.7&71.5&79.5&79.3&79.0\\
\end{tabular}
\captionof{table}{XNLI zero-shot results (Accuracy) for each language}
\label{tab:xnli}
\end{minipage}

\begin{table*}
\centering
\begin{tabular}{lcccccccccccc}
Model&en&ar&de&el&es&hi\\ \hline
mBERT&84.8/72.9&62.6/46.0&72.5/56.8&64.4/47.1&75.3/56.3&58.6/45.1\\
distilmBERT&78.0/65.9&44.6/28.3&57.6/41.0&37.6/21.2&60.5/40.0&34.9/20.5\\
XLM-mlm&77.2/64.5&59.9/43.2&66.0/50.4&57.8/39.5&67.7/49.8&47.5/33.0\\
XLM-tlm&82.5/70.4&68.1/51.6&73.7/57.6&69.5/51.2&77.1/59.2&65.6/50.2\\
XLM-100&84.6/73.4&67.6/50.3&73.6/58.3&63.9/45.1&77.3/59.1&60.2/44.5\\
XLMR-b&83.3/72.4&65.0/47.1&73.4/57.6&71.9/54.5&75.5/57.1&68.3/50.9\\
XLMR-l&86.8/75.5&74.1/55.6&79.5/62.6&79.8/61.4&82.0/62.3&75.4/58.6\\
Model&ru&th&tr&vi&zh&AVG\\ \hline
mBERT&71.4/54.9&43.3/34.4&54.8/40.8&68.1/48.9&58.3/48.2&62.9/47.8\\
distilmBERT&58.9/40.2&20.9/13.9&37.9/21.8&47.5/28.2&46.9/33.8&44.7/28.9\\
XLM&64.1/47.0&24.9/12.4&50.2/34.6&60.3/41.3&39.8/30.1&53.8/38.1\\
XLM-tlm&72.6/55.3&33.3/21.9&65.0/47.5&71.8/51.3&53.4/43.8&65.0/48.9\\
XLM-100&73.7/57.6&22.4/13.6&66.7/49.9&73.9/54.8&54.1/44.5&63.3/47.8\\
XLMR-b&73.3/56.9&67.1/55.5&67.5/50.4&73.0/53.4&51.6/41.7&68.7/52.5\\
XLMR-l&79.4/62.9&73.7/62.6&74.7/58.5&79.4/59.4&55.5/46.7&75.4/59.1\\
\end{tabular}
\captionof{table}{XQuAD results (F$_1$/Exact-match) for each language}
\label{tab:xquad}
\end{table*}

\end{document}